**TITLE: Concurrent validity of computer-vision artificial intelligence player tracking software using broadcast footage**

Zachary L. Crang[a], Rich D. Johnston[a,c,d], Katie L. Mills[e,f], Johsan Billingham[e], Sam Robertson[g], Michael H. Cole[a], Jonathon Weakley[a,c,d], Adam Hewitt[a] and, Grant M. Duthie[bc]

[a] School of Behavioural and Health Sciences, Australian Catholic University, Brisbane, QLD

[b] School of Behavioural and Health Sciences, Australian Catholic University, Strathfield, NSW

[c] Sports Performance, Recovery, Injury and New Technologies (SPRINT) Research Centre, Australian Catholic University, Brisbane, QLD

[d] Carnegie Applied Rugby Research (CARR) Centre, School of Sport, Leeds Beckett University, Leeds, United Kingdom

[e] Fédération Internationale de Football Association (FIFA), Zurich, Switzerland

[f] Sports Engineering Research Group, Sheffield Hallam University, United Kingdom

[g] Institute for Health & Sport, Victoria University, Melbourne, Australia

Corresponding author:

Zachary Crang

Email: Zachary.crang@myacu.edu.au

Phone: +61 420687562

All procedures were approved by a Human Research Ethics Committee (ID: 2023-3480N)

# Concurrent validity of computer-vision artificial intelligence player tracking software using broadcast footage

## Abstract

This study aimed to: (1) understand whether commercially available computer-vision and artificial intelligence (AI) player tracking software can accurately measure player position, speed and distance using broadcast footage and (2) determine the impact of camera feed and resolution on accuracy.

Data were obtained from one match at the 2022 Qatar Fédération Internationale de Football Association (FIFA) World Cup. Tactical, programme and camera 1 feeds were used. Three commercial tracking providers that use computer-vision and AI participated. Providers analysed instantaneous position (x, y coordinates) and speed ($m{\cdot}s^{-1}$) of each player. Their data were compared with a high-definition multi-camera tracking system (TRACAB Gen 5). Root mean square error (RMSE) and mean bias were calculated.

Position RMSE ranged from 1.68 to 16.39 m, while speed RMSE ranged from 0.34 to 2.38 $m{\cdot}s^{-1}$. Total match distance mean bias ranged from -1745 m (−21.8%) to 1945 m (24.3%) across providers.

Computer-vision and AI player tracking software offer the ability to track players with fair precision when players are detected by the software. Providers should use a tactical feed when tracking position and speed, which will maximise player detection, improving accuracy. Both 720p and 1080p resolutions are suitable, assuming appropriate computer-vision and AI models are implemented.

## Introduction

The quantification and interpretation of a player's match or training activities, often termed external load, is widespread in team sports. It is common across various sports to report both aggregated measures (e.g., total distance) as well as discrete phases of play such as the mean peak speed over a 5-minute period [1-3]. This match activity data is used to inform specific training prescription, while training loads are monitored live and retrospectively to ensure the intended training outcomes are achieved. The quantification of match-demands therefore needs to be performed with accurate tracking systems that can also capture enough of the population to provide representative data [4].

Within professional sport, both optical tracking and wearable microtechnology devices are widely used to measure player activities during matches. Optical tracking has developed from notational analysis [5, 6] to semi-automated vision-based tracking systems [7, 8], before automated systems that use computer-based image processing techniques evolved [9-11]. Whilst these systems can provide a high level of accuracy [12], they all require human intervention [10, 11]. Further, optical tracking systems require several fixed or specific camera angles to operate, meaning they are generally only suitable for stadiums with appropriate infrastructure. Wearable microtechnology, which includes global navigation satellite system (GNSS) and local positioning systems (LPS) can circumvent some of the issues with optical tracking. These GNSS devices, allow practitioners to easily quantify the locomotive match-demands of players [2]. Whilst their portability is a strength, the requirement of players to wear the device in a vest or within the playing jersey may impact their use due to regulations of the sport or player preferences.

With developments in computer-vision and artificial intelligence (AI) software, players can be tracked with extremely limited human intervention, using video footage of the match. To collect the footage, single (e.g., wide-angle lens) cameras [13-16] placed on the half-way line, multiple fixed cameras [17], and broadcast acquired footage have been used [15, 18, 19]. Factors that may influence the ability of computer-vision software to accurately detect and track players include occlusion [20], misidentification of players [17], and video resolution [16]. Occlusions can occur where players are gathered closely (e.g., set pieces), misidentification may result when players are similar in appearance (e.g., body shape, boot colour). Further, algorithms using video files with better resolution (2.5k vs. 1080p) were generally more accurate at detecting and tracking players [16]. Given there is no equipment or set-up required by teams when using broadcast footage for player tracking compared to the aforementioned methods, it emerges as the most practical approach. However, the agreement between the outcomes derived from broadcast footage and the other previously established systems is not fully understood.

To the authors' knowledge, no current study has investigated the accuracy of computer-vision and AI software that use broadcast footage to track players position (x, y co-ordinates) and speed. Given that there is likely to be a wide range of proprietary methods that the various providers use to collect, process, and generate the data, there may be substantial bandwidth in data accuracy. Consequently, the aim of this study was to: (1) understand whether commercially available computer-vision and AI player tracking software can accurately measure player position, speed and distance covered using broadcast footage; (2) determine the impact of camera feed and video resolution on accuracy.

## Methods

Data were collected during a single group stage match of the 2022 Qatar Fédération Internationale de Football Association (FIFA) World Cup tournament.

The match was filmed by the television broadcasters using multiple fixed video cameras positioned around the field of play at 50 frames per second (fps) and stored at 25 fps. The tactical, programme and camera 1 video feeds were obtained from the FIFA data platform in MPEG-4 Part-14 (.mp4) file format. The tactical feed is a wide-view angle of the pitch, captured from a fixed camera positioned on the top tier of the grandstand, at the half-way line. The purpose of this camera is to maintain 20 outfield players in shot. The programme feed consists of multiple different camera angles (e.g. tactical, end-on) and is simply what is televised to the public (e.g., including graphics and replays). The camera 1 feed (Figure 1B) captures a slightly tighter field of view compared to the tactical feed. It makes up the majority of the PGM feed, except it does not have any graphical overlay (e.g., scoreboard), replays or cut to different camera angles.

Figure 1 shows an example of the programme, camera 1 and tactical feeds, at the same moment in time. These .mp4 files were downloaded in two different video resolutions; 720p and 1080p. As such, a total of six .mp4 files were provided to the tracking providers to run their analyses.

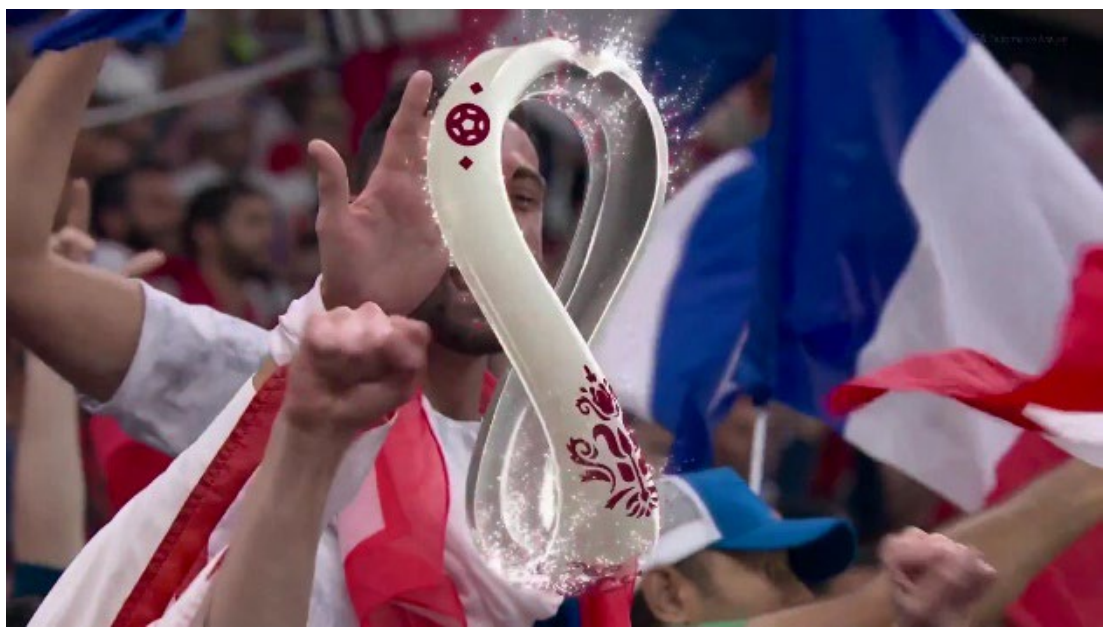

**Figure 1A** Programme feed

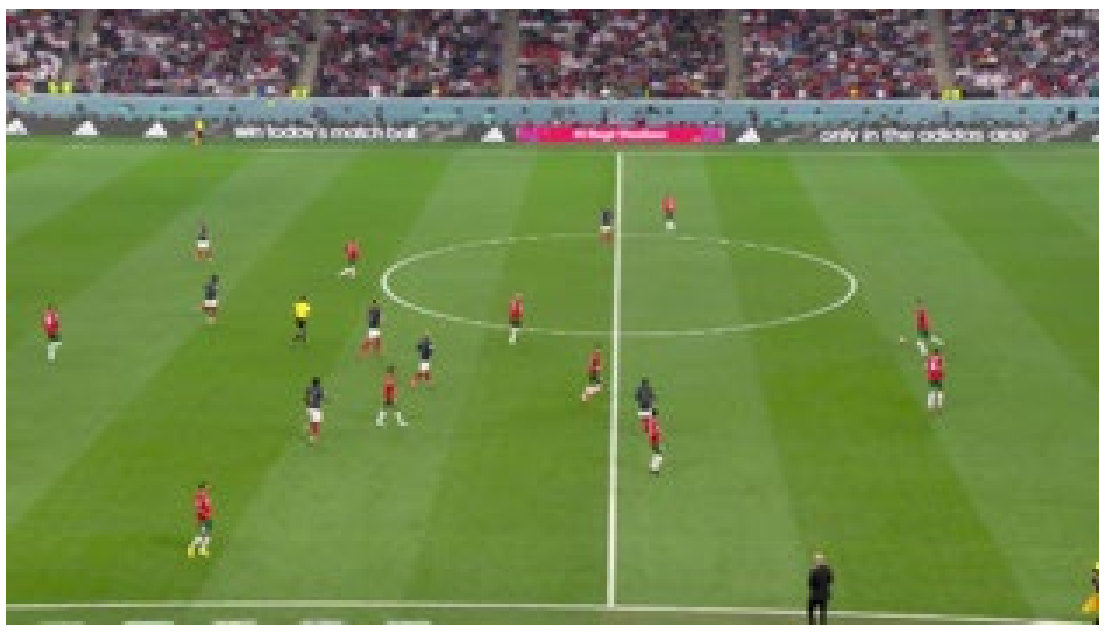

**Figure 1B** Camera 1 feed

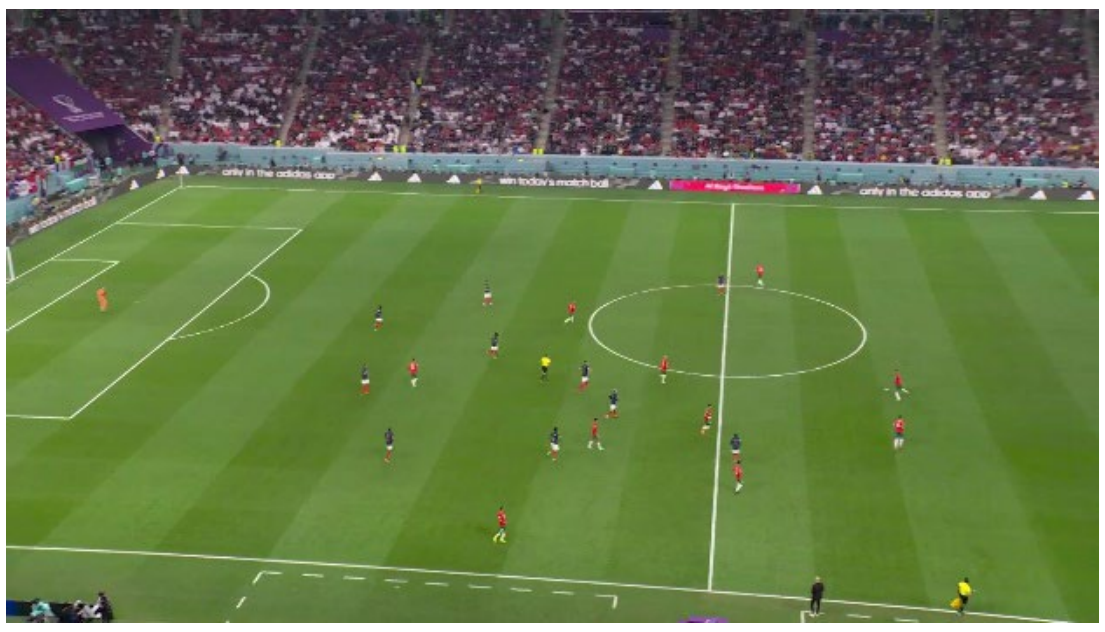

**Figure 1C** Tactical feed

Only players who were on the field of play, including those who entered as substitutes, were tracked for analysis. Given their differing activity profiles to other positions, goalkeepers were removed from the analysis, resulting in a total of 27 players [21]. The match consisted of a 1st half (48 min) and second half (50 min), with no extra time.

Commercially available player tracking providers that use computer-vision and AI to generate position and speed data were invited by FIFA to take part in this study, with three providers volunteering. A combination of undisclosed computer-vision and AI techniques were used to track the overall position (x, y coordinates) and speed ($m \cdot s^{-1}$) of each player at 25 fps. Upon completion, the providers delivered the tracking data in comma-separated value (.csv) files in their raw sampling frequency, where each row provided an observation of player speed and position at 25 Hz. The datasets provided also included detail regarding whether the outcomes were derived when a player was detected or undetected by the provider's software. An example of when the software may not detect the player is when they are positioned outside of the camera's field of view or when a player is obscured by another player. The software understands when a player is on or off the field. Therefore, when processing the image, if a player is expected to be present but is not detected by the software, the embedded AI model is used to generate the player's position and speed data. In this study, the term 'detected' or 'undetected' refers to whether the player was identified by the computer-vision software, while 'visible' indicates that the player was within the camera's field of view.

To establish the concurrent validity, providers were compared against a high-definition multi-camera optical tracking system (TRACAB Gen 5, ChyronHego, New York, USA) capturing data at 25 Hz. TRACAB is a fixed camera system using 12 cameras elevated within the stadium infrastructure. This system has strong accuracy for measures of position (RMSE = 0.08 m) and speed (RMSE = 0.08 $m \cdot s^{-1}$) compared to 3D motion capture [12].

For consistency, the speed from each provider and the TRACAB system were filtered using a $4^{th}$ order 1 Hz low-pass Butterworth filter [22]. Individual player tracking data from each provider for each video resolution were temporally aligned by phase shifting the players' speeds within the TRACAB data to establish the smallest RMSE. The most common phase shift (mode) was then applied to all files. The two datasets were then spatially aligned by rotating the providers' tracking data throughout 360 degrees in 0.1-degree increments to establish the smallest mean error in XY position. Further, the X and Y coordinates were spatially shifted to overlay each other. The most common rotation and shift was found across all files and then applied. The accuracy of the providers to measure overall position and speed was assessed. In addition, the total distance across the entire match was also examined to provide an understanding of how position and speed error influence aggregate data. Total distance was calculated for each player by multiplying their change in speed by change in time.

**Statistical Analysis**

The statistical analyses were performed in RStudio (version 12.1; Posit, Boston, MA) using the R programming language (version 4.3.3, R Foundation for Statistical Computing, Vienna, Austria).

First, to determine the agreement between the different providers and TRACAB, mean bias and the limits of agreement (LOA) were estimated from linear mixed effects (position and speed) and linear models (total distance) built using the *lmerTest::lmer* and *stats::lm* functions. Separate models were built for speed, position and total distance, for each camera (Programme, Camera 1, Tactical) and video resolution (720p, 1080p); all models were fit using restricted estimated maximum likelihood (REML). In each model, the error (i.e. difference to TRACAB) was used as the outcome variable, with the corresponding variable from TRACAB used as a fixed effect. Player ID was included as a random intercept term for the mixed models [23]. The models were run using three different datasets: one including only data where the players were detected by the software, one with only data where they were undetected, and a combined dataset that included all data. From each model, an appropriately weighted mean bias ± 95% limits of agreement (LOA) wase extracted [23, 24]. The root mean square error (RMSE) was separately calculated for position and speed to quantify absolute error. For the purpose of this study, the RMSE for position and speed was interpreted based on their intended use and practical context. Position was considered relative to its use in spatial tracking (e.g., distance between players) rather than other instances like offside decisions. Since the primary objective of these providers is to capture overall player positioning and movement, rather than precise measurements for rule enforcement, an accuracy of ~ 1m was considered suitable. Speed was considered relative to the range of speeds seen in team sports (e.g., 0 to 10 $m \cdot s^{-1}$), with RMSE ~ 10% (~ 1 $m \cdot s^{-1}$) of the maximum speed deemed suitable. These overall position (~ 1 m) and speed (~ 1 $m \cdot s^{-1}$) accuracy thresholds are aligned with current industry standards [25].

$$RMSE = \sqrt{\frac{\sum^{N} (Manufacturer\ Speed - TRACAB\ Speed)^2}{N}}$$

Where *N* is the number of observations in the raw data.

Second, to establish the influence of provider, camera feed (e.g., programme vs. tactical) and video resolution (720p vs. 1080p) on the accuracy of speed, position and total distance, linear mixed models were fit using REML. The RMSE for position and speed, as well as mean bias for total distance were used as the outcome variables, with provider, camera feed and video resolution used as fixed effects in a three-way interaction; player ID was incorporated as a random intercept term. The main effects from each model were extracted using the *stats::anova* function. Where significant main effects were observed, post hoc tests were performed using the *emmeans*::*emmeans* function with a Tukey adjustment applied to account for multiple comparisons. Data are presented as mean ± SD; statistical significance was set at $p < 0.05$.

## Results

### Data Analysed and Detected

The number of datapoints analysed by each provider and percentage within each configuration is outlined in Table 1.

### Positional accuracy

For positional accuracy (Table 2), there were significant main effects for provider ($p<0.001$), camera feed ($p<0.001$), and video resolution ($p<0.001$). The configuration that produced the best accuracy was different across providers. The best accuracy for position was seen for Provider 2 using the 720p tactical feed (RMSE = 1.68 m; mean bias [LOA] = 0.68 m [-2.35 to 3.71 m]). The best for Provider 1 was the 1080p programme feed (RMSE = 3.14 m; mean bias [LOA] = 1.51 m [-3.81 to 6.83 m]), while for Provider 3, the best accuracy was seen for the 1080p camera 1 feed (RMSE = 6.24 m; mean bias [LOA]

= 2.66 m [-8.37 to 13.69 m]). The location of the positional error for the 1080p tactical feed across the 3 providers is illustrated in Figure 2A, 2C and 2E.

Across configurations of camera feed and resolution, Provider 1 had the most consistent accuracy for position compared to TRACAB (RMSE = 3.10 to 6.29 m), with significantly better ($p<0.001$) accuracy for the programme feed and tactical feed compared to the camera 1 feed. The accuracy of Providers 2 and 3 was more variable across configurations, with several significant differences to TRACAB observed (Table 2). Specifically, for Provider 2, there were larger errors for the programme feed at 720p compared to the camera 1 ($p<0.001$) and tactical feed ($p<0.001$). The camera 1 feed was also worse compared to the tactical feed ($p<0.001$). Similarly, at 1080p, the tactical feed was significantly more accurate compared to the camera 1 feed ($p<0.001$) and the programme feed ($p<0.001$). For Provider 3, their tactical feed was significantly worse at 720p compared to the programme ($p<0.001$) and camera 1 ($p<0.001$) feeds; this configuration was also significantly worse than the tactical feed at 1080p ($p<0.001$).

Provider 1 had suitable accuracy when they detected the player from the programme feed (RMSE = 1.07 to 1.14 m). Provider 2 had suitable accuracy across all configurations when the player was detected (RMSE of <1.13 m). Provider 3's accuracy was improved for detected data, although the error was still larger than the other providers. Providers 1 and 2 had the best accuracy for undetected data (RMSE = 4.62 to 12.23 m).

**Speed Accuracy**

For speed accuracy (Table 3), there were significant main effects of provider ($p<0.001$), camera feed ($p<0.001$), and video resolution ($p<0.001$). The configuration that produced the best accuracy was different across providers. The best accuracy for speed was seen by Provider 2 using the 1080p tactical feed (RMSE = 0.34 $m{\cdot}s^{-1}$; mean bias [LOA] = 0.02 $m{\cdot}s^{-1}$ [-0.63 to 0.67 $m{\cdot}s^{-1}$]). The best for Provider 1 was using the 1080p tactical feed (RMSE = 0.39 $m{\cdot}s^{-1}$; mean bias [LOA] = -0.06 $m{\cdot}s^{-1}$ [-0.80 to 0.68 $m \cdot s^{-1}$]) while for Provider 3, the best was for the 1080p camera 1 feed (RMSE = 1.37 $m \cdot s^{-1}$; mean bias [LOA] = 0.26 $m \cdot s^{-1}$ [-2.32 to 2.84 $m \cdot s^{-1}$]). The location of the speed error for the 1080p tactical feed across the 3 providers is illustrated in Figure 2B, 2D and 2F.

Provider 1 had the most consistent accuracy for speed compared to TRACAB (RMSE = 0.4 to 0.6 $m \cdot s^{-1}$). The accuracy of Providers 2 and 3 was more variable across configurations, with some significant differences observed (Table 3). Specifically, for Provider 2, there were larger errors for the programme feed compared to the tactical feed (p<0.001). For Provider 3 tactical feed, the 720p resolution was significantly poorer than the 1080p (p<0.001).

Providers 1 and 2 had the best accuracy when they detected the player from the footage (RMSE: Provider 1 = <0.35 $m \cdot s^{-1}$; Provider 2 = <0.46 $m \cdot s^{-1}$). Provider 3's accuracy was also improved for detected data, although the errors were still larger than the other providers. Providers 1 and 2 had the best accuracy for undetected data (RMSE = 0.78 to 1.13 $m \cdot s^{-1}$).

**Total Distance Accuracy**

The average total distance reported by TRACAB for each player was 7997 ± 3297 m.

For total distance accuracy (Table 4), there were significant main effects of provider (p<0.001) and camera feed (p<0.001). The best accuracy for total distance offered by Provider 2 was derived from the camera 1 1080p feed (mean bias [LOA] = -59 m [-424 to 306 m]). The best accuracy for Provider 1 was using the tactical 1080p feed (mean bias [LOA] = -271 m [-452 to -90 m]), while for Provider 3, it was using the camera 1 1080p feed (mean bias [LOA] = 1163 m [2005 to 321 m]).

Accuracy varied for Providers 1 and 2 across camera feeds, with the programme feed significantly poorer than the camera 1 and tactical feeds (p<0.001). Similarly, for Provider 3, accuracy was significantly reduced for the 1080p programme compared to the camera 1 and tactical feeds (p<0.001).

All providers had suitable accuracy when players were detected regardless of camera feed (mean bias = -298 to 471 m). When players were undetected, accuracy was compromised for Provider 3 across all camera feeds (mean bias = 804 to 1794 m). This was consistent for Providers 2 and 3 using the programme feed (mean bias = -1745 to -750 m). However, using the tactical and camera 1 feeds resulted in improved accuracy (mean bias = -236 to 84 m).

** INSERT TABLE 1, TABLE 2, TABLE 3 NEAR HERE **

** INSERT FIGURE 2 NEAR HERE **

** INSERT TABLE 4 NEAR HERE **

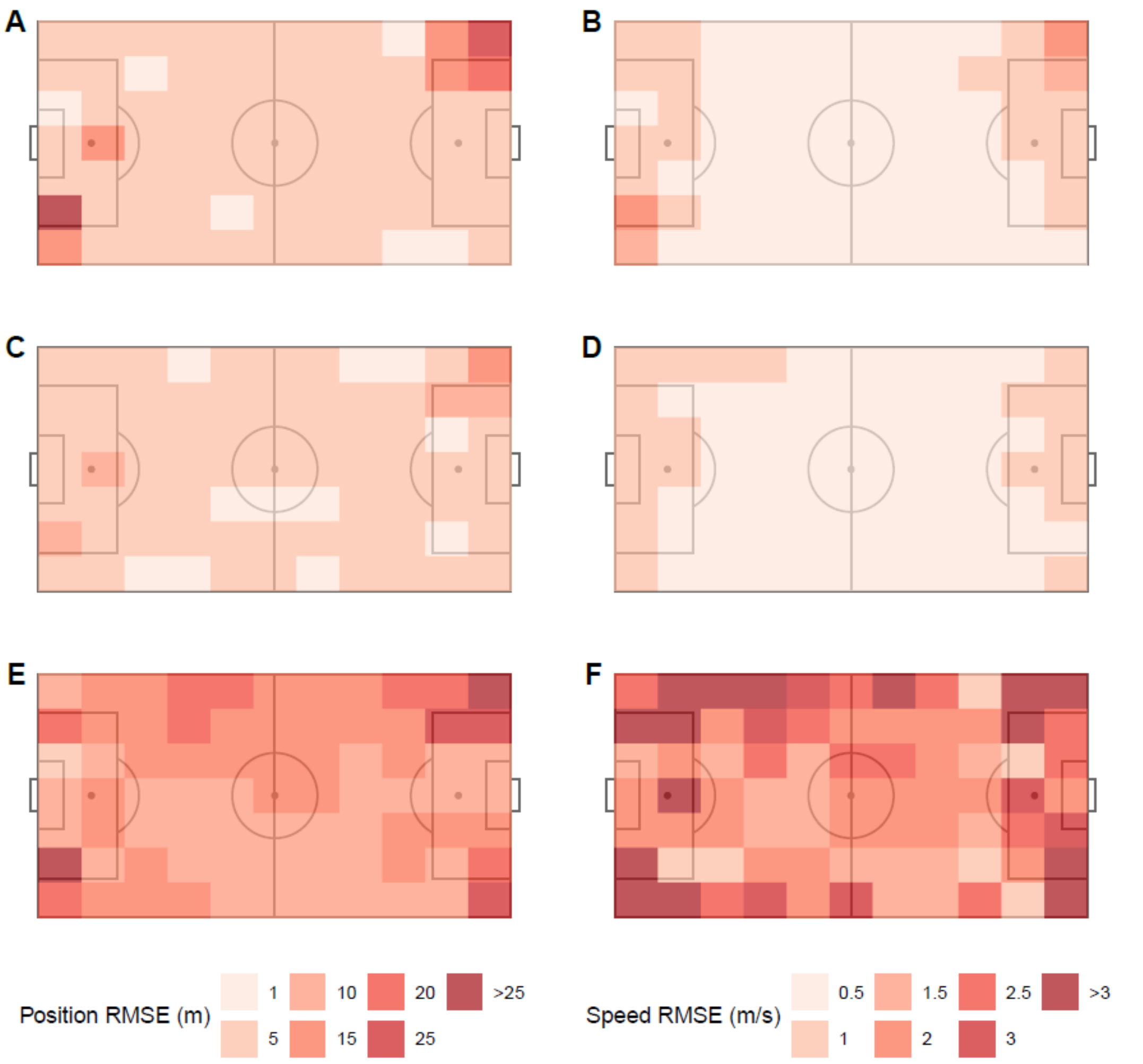

**Figure 2** Location of position (A, C, E) and speed (B, D, F) error for providers 1 (A, B), 2 (C, D) and 3 (E, F)

**Discussion**

The aim of this study was to: (1) understand whether commercially available computer-vision and artificial intelligence (AI) player tracking software can accurately measure player position, speed and distance covered using broadcast footage and (2) determine the impact of camera feed and video resolution on accuracy. It should be made clear that the study's aim was not to appraise the validity of each specific provider, but rather to develop an understanding of how viable computer-vision and AI is for tracking team sport players during competition. The main findings of this study show that computer-vision and AI tracking software can offer suitable validity for measuring speed and player position, but it is influenced by the processing techniques selected by the provider, camera feed and video resolution. Moreover, the detection of the player is crucial for accurate tracking to occur. It is likely that this technology will continue to improve as AI techniques and computational power develops. This study builds on previous research [15, 18, 19], which has demonstrated that players can be accurately detected using broadcast-acquired footage but, to date, has not evaluated the accuracy of tracking their position and speed. Overall, this study shows promise in the use of computer-vision and AI player tracking. Further research is warranted to develop data processing, computer-vision and AI standards for which providers can adhere to and maximise the quality of the data, as well as develop a better understanding of what this technology can be used for.

Overall position accuracy is not currently suitable in the context of spatial tracking (RMSE = 1.68 to 16.39 m). However, Providers 1 and 2 showed that, across multiple configurations, accuracy can be suitable in this context when the player is detected (RMSE = 0.44 to 1.14 m). Therefore, in isolated moments when players are detected by the software (e.g., set play or goal), practitioners could use this data for tactical analysis (e.g., spatial tracking) [26]. Providers 1 and 2 showed suitable speed accuracy across all configurations. For example, a RMSE of 0.34 $m \cdot s^{-1}$ is only 3.4% of the peak speed (10 $m \cdot s^{-1}$) likely to be observed in team sport. While this may be considered suitable, it is not as accurate as GNSS [27]. Overall, practitioners can track player speed using all camera feeds and resolutions. This flexibility is essential for the democratisation of tracking technologies, enabling practitioners who may not have access to all feeds to obtain accurate data with the available resources. However, where possible, the tactical 1080p feed should be used for tracking speeds.

Across providers, there was no clear camera feed or video resolution configuration that consistently produced the best accuracy when measuring position or speed, highlighting that the provider has a significant influence on validity. This is unsurprising given the numerous steps involved in processing the data to track the players (e.g. calibration, filter type, machine learning approaches used) will differ between providers. Prior to assessing player movements, the playing area must first be calibrated to understand the position and dimensions of the pitch relative to the camera. This may involve various homography techniques which are commonly used in the context of computer-vision. Second, the software must then be able to detect individual players, which will typically rely on defining attributes such as boot colour and playing shirt number. Third, given the number of machine learning algorithms available (e.g., You Only Look Once, Convolutional Neural Networks), provider-developed AI models will differ as well as the way in which model hyperparameters are tuned. Fourth, several computer-vision and AI techniques allow for multi-object tracking during match-play, whilst players are in and out of the camera field of view. Like GNSS providers, this intellectual property is not disclosed to the end-user, and it is therefore difficult to determine the best-practice AI and data processing methods to implement in this ever-evolving field. Given all, except Provider 1, recorded poor validity when using the programme feed to measure position, it would appear they implement AI techniques superior to the other providers. The major difference between providers using this feed appears to come from when players were undetected, with Provider 1 reporting a RMSE of 4.6 m for position during undetected frames, as opposed to 10.0 to 11.6 m for the other providers. This suggests that Provider 1 implemented a superior AI extrapolation model that can better determine the position of a player when outside of the camera's field of view. This appears to be complex however, and may depend on the camera feed used. Overall, the providers are not currently capable of accurately estimating position when players are not detected by the software (e.g., outside of the field of view of the camera). Speed however is deemed suitable for Providers 1 and 2 and can be used by practitioners, though is still significantly poorer compared to when players were detected. Future research should focus on improving the extrapolation models used by the providers to enhance overall accuracy.

Player detection is key in maximising accuracy. This is supported by Providers 1 and 2 who have superior accuracy for position (RMSE = 0.44 to 2.23 m) and speed (RMSE = 0.20 to 0.46 $m \cdot s^{-1}$) from frames when a player was detected compared to when they were not. Thus, most of the overall error can be attributed to when the player is undetected, further highlighting that refining the extrapolation models used to estimate position and speed is pivotal to enhancing overall accuracy. To improve player detection, it appears that minimising frames where players are out of the cameras field of view is important. For example, there is a significant improvement in position validity for Provider 2 when using the tactical feed compared to the programme and camera 1 feeds. The tactical feed provides an elevated and wide angle of the field, increasing the field of view (i.e., increasing player visibility) and limiting occlusion [28]. Therefore, the computer-vision tracking software can detect the player for a greater number of frames (89 to 96% detection) compared to the other feeds (36 to 64% detected), again relying much less on the extrapolation to estimate position when the player is not detected. Ensuring practitioners can determine whether a player was detected may be useful for tactically analysing isolated periods of play (e.g., opposition set pieces) to gain greater confidence in the data. If it is found that the player was not detected, this data should not be used in such contexts. Despite showing a significant effect of video resolution on validity, there does not appear to be any practically meaningful changes between video resolution configurations.

While this research focuses on the ability of the software to determine position and speed, it is important to consider the influence it has on commonly reported metrics such as total distance covered that is derived from these metrics. It appears that Providers 1 and 2 offered suitable measures of total distance when using the camera 1 (percentage error = -0.74 to -5.73%) and tactical feeds (percentage error = -1.20 to 3.78%), while Provider 3 reported poor validity (percentage error = 14.54 to 24.32%). Once again, distance accuracy appears linked to player detection, with player detection greater for the camera 1 and tactical feeds compared to the programme feed. This was highlighted by Provider 2, as most of the error came from situations where a player was undetected. It is important to note that while the accuracy of position and speed appeared suitable for Provider 1 using the programme feed, and at times better than other configurations, when the data were aggregated to calculate total distance, there was a large difference compared to TRACAB. This potentially could be explained by the direction of the error, with even small errors in same direction capable of accumulating and magnifying the discrepancy in total distance. In contrast, other configurations may present less accuracy at individual data points, but if this error varies in direction, then the influence on total distance is decreased.

Overall, it is recommended that providers use a tactical feed with 720p or 1080p video resolution when tracking players' position and speed. This method will maximise the number of video frames the players are visible (i.e., in the camera's field of view) and detected for, improving accuracy. This however is reliant upon the provider implementing the correct computer-vision and AI model.

A limitation of this study was that the data was collected from a single match, stadium and broadcaster in what could be considered prime conditions. In turn, stadium design, camera angles and jersey colour may vary at other stadiums which could influence the validity of the software. For example, a stadium with a tactical camera that is at a lower elevation (e.g. narrower field of view) may introduce more undetected data points. Similarly, camera operators or broadcasters may vary in the way that they film the match (e.g. holding a wide-shot vs zooming in and out), which could also result in more undetected data points. Nonetheless, this study shows that players can be accurately tracked using broadcast footage if the provider implements a suitable computer-vision and AI model, and the players are detected by the tracking software.

## Conclusions

Players can be tracked with computer-vision and AI software that uses video, though accuracy is dependent on the provider selecting a suitable computer-vision and AI model, as well as camera feed and video resolution. It is important to maximise the number of frames in which a player is detected, which is achieved by increasing their visibility by using a tactical feed (i.e., wide-view elevated camera). This is highlighted in this study with player detection much greater using the tactical feed (89 to 96%) compared to the programme and camera 1 feeds (36 to 64%). The AI techniques implemented by the provider appears to have an influence, with some providers showing superior validity compared to others when a player is not detected by the software. Regardless, providers do not have suitable accuracy when the player is undetected by the software for measures of position. Validity is suitable however for measures of speed. Regardless, future research should focus on refining the extrapolation methods implemented by the providers to improve overall accuracy. While there was a significant effect of video resolution on validity, there was no practically meaningful differences between configurations. It is important to also consider the influence of the software's position and speed accuracy on derived variables such as aggregated total distance. Overall, it is recommended providers use a tactical feed, while 720p or 1080p is an adequate video resolution to track players. Consumers should be aware that validity may change between providers, given they may use different computer-vision and AI models.

**Statements and Declarations**

No funding was received for conducting this study.

**Table 1.** The number (% total) of data points detected by each provider using the Programme (PGM), Camera 1 and Tactical feeds at the 720p and 1080p resolutions

| | | Datapoints (%) | | | | | |
|---|---|---|---|---|---|---|---|
| | | **720p** | | | **1080p** | | |
| | | **PGM** | **Cam 1** | **Tactical** | **PGM** | **Cam 1** | **Tactical** |
| **Provider 1** | Detected | 1,236,404 (42%) | 1,883,411 (63%) | 2,802,310 (94%) | 1,236,589 (42%) | 1,881,219 (63%) | 2,819,394 (95%) |
| | Undetected | 1,732,652 (58%) | 1,092,729 (37%) | 173,723 (6%) | 1,732,467 (58%) | 1,094,921 (37%) | 156,639 (5%) |
| | Overall | 2,969,056 | 2,976,140 | 2,976,033 | 2,969,056 | 2,976,140 | 2,976,033 |
| **Provider 2** | Detected | 1,078,425 (36%) | 1,661,976 (56%) | 2,647,867 (89%) | 1,125,202 (38%) | 1,726,834 (58%) | 2,680,068 (90%) |
| | Undetected | 1,881,935 (64%) | 1,298,482 (44%) | 316,005 (11%) | 1,835,158 (62%) | 1,233,624 (42%) | 283,800 (10%) |
| | Overall | 2,963,872 | 2,960,458 | 2,963,872 | 2,963,872 | 2,960,458 | 2,963,868 |
| **Provider 3** | Detected | 1,203,392 (42%) | 1,845,274 (64%) | 2,775,052 (96%) | 1,202,991 (42%) | 1,846,269 (64%) | 2,757,273 (95%) |
| | Undetected | 1,690,595 (58%) | 1,048,713 (36%) | 119,418 (4%) | 1,690,996 (58%) | 1,048,285 (36%) | 137,197 (5%) |
| | Overall | 2,893,987 | 2,893,987 | 2,894,470 | 2,893,987 | 2,894,554 | 2,894,470 |

PGM; Programme feed

**Table 2.** Concurrent validity of player software tracking systems to measure position (m) during football match-play in comparison to an optical tracking system.

| | | **RMSE (m)** | | | | | | **Mean Bias (m) ± 95% LOA** | | | | | |
|---|---|---|---|---|---|---|---|---|---|---|---|---|---|
| | | **720p** | | | **1080p** | | | **720p** | | | **1080p** | | |
| | | **PGM** | **Cam 1** | **Tactical** | **PGM** | **Cam 1** | **Tactical** | **PGM** | **Cam 1** | **Tactical** | **PGM** | **Cam 1** | **Tactical** |
| **Provider 1** | Detected | 1.14 | 2.23 | 1.79 | 1.07 | 1.69 | 1.68 | 0.53 ± 1.99 | 0.79 ± 4.09 | 0.82 ± 3.27 | 0.54 ± 1.82 | 0.70 ± 3.01 | 0.80 ± 2.94 |
| | Undetected | 4.62 | 9.56 | 10.82 | 4.58 | 9.65 | 12.23 | 2.80 ± 7.24 | 5.29 ± 16.62 | 5.74 ± 17.09 | 2.81 ± 7.13 | 5.14 ± 16.08 | 6.13 ± 20.66 |
| | Overall | 3.14[a] | 6.29[b] | 3.14 | 3.10[a] | 6.00[b] | 3.25 | 1.50 ± 5.42 | 2.40 ± 11.38 | 1.17 ± 5.89 | 1.51 ± 5.32 | 2.31 ± 10.85 | 1.11 ± 6.02 |
| **Provider 2** | Detected | 1.13 | 0.66 | 0.47 | 1.02 | 0.68 | 0.44 | 1.03 ± 0.92 | 0.60 ± 0.59 | 0.41 ± 0.47 | 0.95 ± 0.77 | 0.59 ± 0.70 | 0.37 ± 0.47 |
| | Undetected | 11.56 | 9.27 | 4.96 | 9.57 | 9.08 | 6.19 | 7.75 ± 16.58 | 6.42 ± 13.24 | 2.95 ± 8.15 | 6.57 ± 13.72 | 6.15 ± 13.25 | 3.59 ± 9.66 |
| | Overall | 9.24[ab] | 6.16[b] | 1.68 | 7.57[b] | 5.89[b] | 1.96 | 5.36 ± 14.71 | 3.14 ± 10.42 | 0.68 ± 3.03 | 4.45 ± 12.04 | 2.91 ± 10.08 | 0.72 ± 3.61 |
| **Provider 3** | Detected | 3.62 | 5.03 | 16.24 | 3.05 | 2.49 | 9.95 | 1.48 ± 6.49 | 2.78 ± 8.15 | 8.69 ± 26.75 | 1.37 ± 5.38 | 0.85 ± 4.58 | 3.34 ± 17.61 |
| | Undetected | 10.03 | 10.27 | 19.38 | 10.02 | 9.83 | 15.03 | 6.37 ± 15.11 | 6.53 ± 15.46 | 11.25 ± 26.16 | 6.33 ± 15.17 | 5.96 ± 15.30 | 7.67 ± 23.49 |
| | Overall | 8.01[b] | 7.37[b] | 16.39[$] | 7.91 | 6.24 | 10.25 | 4.31 ± 13.18 | 4.11 ± 11.91 | 8.78 ± 26.70 | 4.24 ± 13.04 | 2.66 ± 11.03 | 3.55 ± 18.05 |

RMSE; root mean square error, LOA; limits of agreement, PGM; Programme feed
a; significant difference ($p < 0.05$) to cam 1 feed for the same provider and resolution
b; significant difference ($p < 0.05$) to tactical camera feed for the same provider and resolution
$; significant difference ($p < 0.05$) between camera resolution for the same provider and camera feed

**Table 3.** Concurrent validity of player software tracking systems to measure speed (m·s$^{-1}$) during football match-play in comparison to an optical tracking system

| | | RMSE (m·s$^{-1}$) | | | | | | Mean Bias (m·s$^{-1}$) ± 95% LOA | | | | | |
|---|---|---|---|---|---|---|---|---|---|---|---|---|---|
| | | 720p | | | 1080p | | | 720p | | | 1080p | | |
| | | PGM | Cam 1 | Tactical | PGM | Cam 1 | Tactical | PGM | Cam 1 | Tactical | PGM | Cam 1 | Tactical |
| **Provider 1** | Detected | 0.35 | 0.34 | 0.35 | 0.35 | 0.32 | 0.32 | -0.10 ± 0.65 | -0.08 ± 0.64 | -0.08 ± 0.66 | -0.10 ± 0.65 | -0.08 ± 0.61 | -0.07 ± 0.61 |
| | Undetected | 0.78 | 0.91 | 1.10 | 0.78 | 0.89 | 1.02 | -0.20 ± 1.34 | -0.16 ± 1.62 | -0.06 ± 1.80 | -0.20 ± 1.34 | -0.16 ± 1.59 | 0.05 ± 1.72 |
| | Overall | 0.57 | 0.61 | 0.43 | 0.57 | 0.60 | 0.39 | -0.14 ± 1.06 | -0.11 ± 1.15 | -0.08 ± 0.81 | -0.14 ± 1.06 | -0.11 ± 1.13 | -0.06 ± 0.74 |
| **Provider 2** | Detected | 0.39 | 0.27 | 0.24 | 0.46 | 0.30 | 0.20 | -0.04 ± 0.76 | -0.004 ± 0.51 | 0.02 ± 0.46 | 0.03 ± 0.90 | 0.01 ± 0.58 | 0.001 ± 0.39 |
| | Undetected | 1.13 | 0.95 | 0.80 | 0.94 | 0.89 | 0.91 | -0.60 ± 1.44 | -0.08 ± 1.47 | 0.31 ± 1.25 | -0.27 ± 1.49 | -0.04 ± 1.41 | 0.18 ± 1.43 |
| | Overall | 0.93$^{b}$ | 0.66 | 0.34 | 0.79$^{b}$ | 0.62 | 0.34 | -0.40 ± 1.56 | -0.04 ± 1.20 | 0.05 ± 0.65 | -0.16 ± 1.46 | -0.01 ± 1.14 | 0.02 ± 0.65 |
| **Provider 3** | Detected | 1.19 | 1.37 | 1.73 | 1.12 | 0.86 | 1.40 | 0.10 ± 2.33 | 0.07 ± 2.61 | 0.11 ± 3.31 | 0.09 ± 2.20 | 0.04 ± 1.68 | 0.09 ± 2.71 |
| | Undetected | 2.02 | 2.27 | 8.21 | 2.00 | 1.97 | 6.75 | 0.70 ± 3.51 | 0.74 ± 3.97 | 3.86 ± 13.39 | 0.68 ± 3.48 | 0.64 ± 3.41 | 2.61 ± 11.28 |
| | Overall | 1.73 | 1.75 | 2.38$^{\$}$ | 1.69 | 1.37 | 2.01 | 0.45 ± 3.16 | 0.31 ± 3.25 | 0.28 ± 4.54 | 0.43 ± 3.10 | 0.26 ± 2.58 | 0.22 ± 3.85 |

RMSE; root mean square error, LOA; limits of agreement, PGM; Programme feed
a; significant difference ($p < 0.05$) to cam 1 feed for the same provider and resolution
b; significant difference ($p < 0.05$) to tactical camera feed for the same provider and resolution
$; significant difference ($p < 0.05$) between camera resolution for the same provider and camera feed

**Table 4.** Concurrent validity of player software tracking systems to measure total distance (m) during football match-play in comparison to an optical tracking system

| | | Mean Bias (m) ± 95% LOA | | | | | |
|---|---|---|---|---|---|---|---|
| | | 720p | | | 1080p | | |
| | | **PGM** | **Cam 1** | **Tactical** | **PGM** | **Cam 1** | **Tactical** |
| **Provider 1** | Detected | -184 ± 62 | -222 ± 93 | -298 ± 140 | -175 ± 63 | -212 ± 82 | -287 ± 109 |
| | Undetected | -1489 ± 381 | -236 ± 353 | -3 ± 109 | -1490 ± 384 | -231 ± 350 | 16 ± 108 |
| | Overall | -1672 ± 398$^{ab}$ | -458 ± 386 | -302 ± 220 | -1665 ± 404$^{ab}$ | -442 ± 380 | -271 ± 181 |
| **Provider 2** | Detected | -56 ± 64 | -9 ± 30 | 66 ± 114 | 44 ± 102 | 15 ± 37 | 12 ± 51 |
| | Undetected | -1689 ± 1107 | -150 ± 456 | 158 ± 136 | -750 ± 401 | -74 ± 351 | 84 ± 152 |
| | Overall | -1745 ± 1099$^{ab\$}$ | -159 ± 469 | 225 ± 191 | -706 ± 426$^{ab}$ | -59 ± 365 | 96 ± 163 |
| **Provider 3** | Detected | 150 ± 267 | 174 ± 488 | 591 ± 2176 | 130 ± 206 | 109 ± 230 | 471 ± 1190 |
| | Undetected | 1794 ± 1106 | 1222 ± 1427 | 804 ± 1984 | 1747 ± 999 | 1054 ± 791 | 713 ± 1955 |
| | Overall | 1945 ± 1303 | 1395 ± 1775 | 1395 ± 4116 | 1877 ± 1124$^{ab}$ | 1163 ± 842 | 1184 ± 3776 |

LOA; limits of agreement, PGM; Programme feed
a; significant difference ($p < 0.05$) to cam 1 feed for the same provider and resolution
b; significant difference ($p < 0.05$) to tactical camera feed for the same provider and resolution
$; significant difference ($p < 0.05$) between camera resolution for the same provider and camera feed